\documentclass[10pt,twocolumn,letterpaper]{article}

\usepackage{cvpr}              % To produce the CAMERA-READY version
\usepackage{makecell}

\definecolor{cvprblue}{rgb}{0.21,0.49,0.74}
\usepackage[pagebackref,breaklinks,colorlinks,allcolors=cvprblue]{hyperref}

\def\paperID{9327} % *** Enter the Paper ID here
\def\confName{CVPR}
\def\confYear{2025}

\title{\textsl{FocusLLaVA}: A Coarse-to-Fine Approach for Efficient and Effective Visual Token Compression}

\author{
Yuke Zhu$^1$\footnotemark[1], Chi Xie$^2$\footnotemark[1], Shuang Liang$^2$, Bo Zheng$^1$, Sheng Guo$^1$\\
{$^1$Mybank, Ant Group\hspace{0.5cm}}
{$^2$Tongji University\hspace{0.5cm}}\\
{\texttt{\small{\{felix.yk,guangyuan,guosheng.guosheng\}@mybank.cn}}\hspace{0.5cm}}
{\texttt{\small{\{chixie,shuangliang\}@tongji.edu.cn}}}
}

\begin{document}

\maketitle

\renewcommand{\thefootnote}{\fnsymbol{footnote}}
\footnotetext[1]{Equal contribution.}

\begin{abstract}
    Recent advances on Multi-modal Large Language Models have demonstrated that high-resolution image input is crucial for model capabilities, especially for fine-grained tasks. However, high-resolution images lead to a quadratic increase in the number of visual tokens input into LLMs, resulting in significant computational costs. Current work develop visual token compression methods to achieve efficiency improvements, often at the expense of performance. We argue that removing visual redundancy can simultaneously improve both efficiency and performance. We build a coarse-to-fine visual token compression method, with a vision-guided sampler for compressing redundant regions with low information density, and a text-guided sampler for selecting visual tokens that are strongly correlated with the user instructions.
    With these two modules, the proposed \textsl{FocusLLaVA} achieves improvements in both efficiency and performance. We validate the effectiveness of our approach on a wide range of evaluation datasets.
\end{abstract}

\section{Introduction}
Recently, the study of Multimodal Large Language Models (MLLMs) has attracted considerable attention from researchers. With a variety of MLLMs like LLaVA~\cite{DBLP:conf/nips/LiuLWL23a_llava,Liu_2024_CVPR_llava1_5} being proposed, significant progress has been made in this field.
One of the major improvement for recent MLLMs is the support for high-resolution images. Early works~\cite{DBLP:conf/nips/LiuLWL23a_llava,DBLP:conf/icml/BLIP2} use a fixed, small input scale, regardless of the original image size. This process inevitably led to the loss of image details, making the model incapable of handling tasks that require fine-grained image understanding.
Recently, supporting high-resolution image input~\cite{Monkey,qwen-vl} has become a common interest in the industry. However, high-resolution images lead to a quadratic increase in the number of visual tokens in the LLM, resulting in increased inference time and higher memory consumption.

To address the problem, numerous studies have focused on reducing the number of visual tokens in high-resolution images. 
However, most of these methods employ a heuristic approach for visual token compression~~\cite{DBLP:journals/corr/m3,zhang2024beyond,arif2024hired,song2024less}. They leverage hand-crafted metrics to filter out tokens, which often at the price of the performance.
Some other works~\cite{DBLP:conf/icml/BLIP2,yu2024texthawk,zhang2024beyond} design query transformers to compress image information into a fixed number of queries, which lead to information loss for detailed images and sometimes a complex training scheme.
Generally, current approach cannot guarantee maintaining model performance while reducing visual tokens.
They focus on achieving a trade-off between faster speed and less performance loss.
% These approaches can be broadly categorized into three categories:using QFormer-like structure~\cite{DBLP:conf/icml/BLIP2,yu2024texthawk,zhang2024beyond}, designing heuristic importance metric~\cite{DBLP:journals/corr/m3,zhang2024beyond,arif2024hired} and borrowing pruning technique from LLM~\cite{DBLP:journals/corr/FastV,DBLP:journals/corr/VTW}.
% The first category involves designing resampling structures similar to QFormer~\cite{DBLP:conf/icml/BLIP2,yu2024texthawk,zhang2024beyond}.
% Such methods typically compress images using a fixed number of tokens, without considering the image's resolution, which inevitably leads to information loss. 
% The second category employs heuristic technique for visual token compression~\cite{DBLP:journals/corr/m3,zhang2024beyond,arif2024hired}. These manually designed approaches cannot guarantee optimal choice and often result in performance degradation. 
% The third category~\cite{DBLP:journals/corr/FastV,DBLP:journals/corr/VTW} is based on token pruning techniques used in large language models (LLMs). These methods generally focus on optimizing inference speed and memory usage without retraining, with less emphasis on improving model performance. 
% Overall, current mainstream approaches mainly work by striking a balance between performance and efficiency.

% In this paper, we argue that removing visual redundancy can simultaneously improve performance and efficiency, as eliminating distractions allows the model to focus more on areas related to the problem and answer. 
In this paper, we propose \textsl{FocusLLaVA}, which removes visual redundancy and simultaneously improves both performance and efficiency.
% We achieve this by utilizing both visual and textual information for selecting high-value visual tokens. 
% The insight behind \textsl{FocusLLaVA} is simple: in vision-language tasks, the information in an image can be categorized into (i) regions like the background that contain a large amount of redundant information, (ii) regions with high information density and rich details, and (iii) regions that are strongly correlated with the user's query. 
% Intuitively, only the third part is necessary, but in reality, it is challenging to precisely locate such region in a lightweight and straightforward manner.
% Therefore, we designed a coarse-to-fine approach to achieve the final goal: 
It uses a coarse-to-fine approach for this target:
it first compress low information density features based on visual information (\textbf{Vision-Guided Sampler}), and then select tokens relevant to the query based on textual instructions (\textbf{Text-Guided Sampler}).
% Initially, we use visual guidance to differentiate regions of varying information density within the image, filtering out redundant information while preserving as much detail as possible. 
% In the first step, we design a module named vision-guided sampler to differentiate and process regions of varying information density within the image. 
% In the second step, we design a module named text-guided sampler to further identify areas strongly associated with the user's instructions.
% his part is called the Text-guided Sampler. In this module, we use the text embeddings from LLM serve as the textual guidance. The visual tokens are selected based on their similarity with the text embeddings.

We make several technical designs for these two modules. 
First,  the \textbf{region-level compression} in vision-guided sampler: it down-samples each local region with multiple scales and select one of them adaptively.
We use regions rather than token as the basic units to enables a multi-scale manner, which is more flexible.
Second, the \textbf{disentangled compression} of the two samplers: the vision-guided sampler is placed in the projector, as it requires only image information and serves as a coarse step for reducing redundant tokens early. In contrast, the text-guided sampler is integrated within the intermediate layers of the LLM, as it necessitates stronger language features to pinpoint areas related to the instructions.
Third, a straightforward, one-step \textbf{training recipe} carefully designed for both modules. For the former, we use an auxiliary balance loss to encourage the vision-guided sampler to explore various scales. For the latter, we design a stochastic training technique to adapt the LLM to token changes. The fine-tuning step is one-stage, as the baseline.

% The two modules are carefully designed to ensure that important tokens are preserved. 
% For the vision-guided sampler, we choose to use local feature blocks, rather than each single token, as the basic unit for processing, so as to support the handling of local regions in a multi-scale manner. 
% This module shares parameters across all feature blocks and predicts the scale by which each feature block needs to be down-sampled. This approach, where the model learns on its own instead of through manual design, ensures that the model can adaptively make better scale choices for each feature block.
% Additionally, during training, we introduce a balance loss to encourage the model to explore different compression scales.
% For the text-guided sampler, we found that only after passing through several layers of LLM can the text embeddings effectively focus on the relevant areas in the image. Based on this observation, the text-guided sampler is embedded within the LLM to achieve precise selection. The text-guided sampler is trained using a stochastic method, which involves randomly selecting different layers and varying the number of visual tokens to enhance the LLM's adaptability to token changes.
% The two additional modules we added do not require extra training stages; they can simply be trained alongside the model during the Instruction Fine Tuning phase.

To validate the effectiveness of \textsl{FocusLLaVA}, we conduct experiments on various mainstream multimodal evaluation benchmarks. The results demonstrates that our approach not only optimizes the inference speed, but also improves performance. With 39\% visual tokens, \textsl{FocusLLaVA} outperforms the baseline method in a wide range of benchmarks.
% Moreover, we carried out extensive comparative experiments to study the individual roles and characteristics of the proposed visual and textual guidance. These components are very helpful for understanding the design of our method and for future work considerations.
In summary, the main contributions of our work can be summarized as follows:
\begin{enumerate}
    \item We propose \textsl{FocusLLaVA}, a novel coarse-to-fine approach for visual token reduction, which leverages the guidance from both visual and textual information.
    \item We carefully designed two modules, the vision-guided sampler and the text-guided sampler, along with a complete training methodology.
    \item We demonstrate both efficiency and performance improvements of \textsl{FocusLLaVA} on broad benchmarks. It shows highly competitive results compared to SOTA MLLMs. Extensive experiments are conducted for better understanding of both visual and textual guidance.
\end{enumerate}

\begin{figure*}[!t] % 注意 * 和位置参数 [!t]
    \centering
    \includegraphics[width=\textwidth]{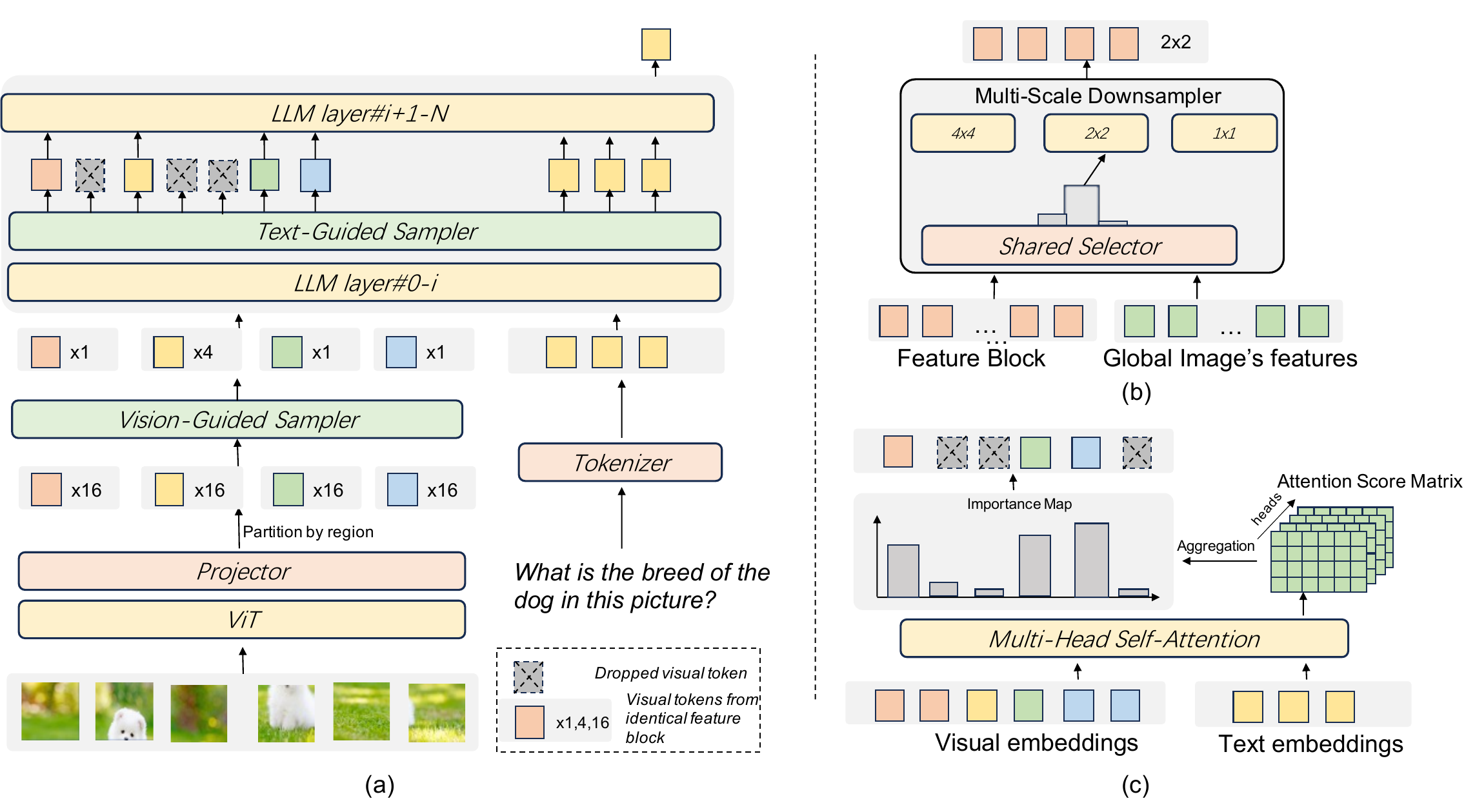} % 修改为你自己的图片文件路径
    \caption{(a) \textbf{Overall structure of \textsl{FocusLLaVA}}. The two core modules are vision-guided sampler and text-guided sampler. The features from each sub-images are first concatenated into a whole and then partitioned by regions, each forming a local feature block. It is then processed by vision-guided sampler.
    (b) \textbf{The structure of vision-guided sampler.}
    It takes a feature block and global image's features as inputs and output the predicted sampling scale for this region.
    (c) \textbf{The structure of text-guided sampler.} It aggregates the multi-head attention scores to form the importance map of visual tokens.}
    \label{fig:main}
\end{figure*}

\section{Related Work}
\textbf{MLLMs}.
Early models like BLIP2~\cite{DBLP:conf/icml/BLIP2} and InstructBLIP~\cite{DBLP:conf/nips/InstructBLIP} designed a Q-Former to bridge encoded visual information into the input space of LLMs. 
These approaches typically require complex training processes to train the image-text alignment module. 
Methods represented by Flamingo~\cite{DBLP:conf/nips/Flamingo} proposed incorporating encoded image information into LLM layers using cross-attention. 
Fuyu8B~\cite{fuyu-8B} entirely discarded the visual encoder, directly inputting image patches into the LLM. 
LLaVA~\cite{DBLP:conf/nips/LiuLWL23a_llava}, on the other hand, uses an MLP layer to directly bridge encoded image information into the LLM's input space, making the model architecture and training process much more straightforward. 
Consequently, many subsequent multimodal large models have made improvements based on LLaVA~\cite{liu2024llavanext, Liu_2024_CVPR_llava1_5, DBLP:journals/corr/ALLaVA, DBLP:journals/corr/MOE-LLaVA, DBLP:journals/corr/MQT-LLAVA}. 
For instance, LLaVA 1.5~\cite{Liu_2024_CVPR_llava1_5} optimized data quality, while LLaVA-Next~\cite{liu2024llavanext} introduced adaptive image segmentation techniques to support high-resolution images.

\noindent\textbf{High-resolution MLLM.}
Recently, various MLLMs have adopted high-resolution images as input to capture fine-grained image information. 
In the early stages, most MLLMs use a fixed size of 224 for their inputs. 
LLaVA-1.5~\cite{Liu_2024_CVPR_llava1_5} and BLiVA~\cite{hu2024bliva} increased the image size to 336 to achieve better performance. 
Qwen-VL~\cite{qwen-vl} expanded the size further to 448. 
It first trained with a fixed image scale of 224 and then fine-tuned by increasing the resolution to 448. 
Vary~\cite{DBLP:journals/corr/vary} and Mini-Gemini~\cite{DBLP:journals/corr/Mini-Gemini} additionally introduced a Vision Encoder specifically for high-resolution images. 
SPHINX~\cite{DBLP:journals/corr/SPHINX}, Monkey~\cite{Monkey}, and LLaVA-UHD~\cite{xu2024llava-uhd} resize the images to a fixed resolution and then split them into several patches. 
These patches are individually encoded before being fed to the LLM. 
Furthermore, LLaVA-NeXT~\cite{Open-LLaVA-NeXT} employs a series of predefined resolutions, first matching the input image to the closest resolution, and then segmenting it into sub-images.
Qwen2-VL~\cite{wang2024qwen2vl} directly uses the original resolution of the image and encodes it into dynamically variable-length visual tokens by modifying the structure of ViT~\cite{dosovitskiy2021ViT}.
LLaVA-HR~\cite{luo2024feast}, LLaVA-M3~\cite{DBLP:journals/corr/m3}, and Dragonfly~\cite{chen2024dragonfly} advocate for the use of multi-scale information to enhance MLLM capabilities. 
They resize the original images into multiple resolutions, each encoded individually, and then feed them into the LLM. 
This approach leverages the benefits of different scales, providing a more comprehensive representation of the input data.

\noindent\textbf{Visual token compression.}
In the field of MLLM, the works for visual token compression can be broadly categorized into three types. The first type~\cite{zhang2024beyond,yu2024texthawk, yu2024texthawk2} employs a QFormer-like~\cite{DBLP:conf/icml/BLIP2,DBLP:conf/nips/InstructBLIP} structure to compress visual tokens. 
% The original QFormer utilizes a set of learnable query embeddings to integrate visual and textual information. 
However, they heavily relies on the quality of alignment between image and text, necessitating a complex training process for alignment. Subsequent works, such as QwenVL~\cite{qwen-vl}, employs a single layer of cross-attention to replace QFormer, thereby reducing the training complexity. However, it still uses a fixed number of queries for image information compression, which lead to information loss, especially in tasks requiring fine-grained information.
The second category~\cite{DBLP:journals/corr/LLaVA-PruMerge,chen2024dragonfly, zhang2024tokenlevel, cao2024madtp,DBLP:journals/corr/textmonkey,DBLP:journals/corr/Otterhd, song2024less} leverages manually designed techniques for compressing image information. For instance, LLaVA-PruMerge~\cite{DBLP:journals/corr/LLaVA-PruMerge} dynamically identifies and retains the most critical visual tokens, then merges similar ones through clustering. 
% Dragonfly~\cite{chen2024dragonfly} uses mean-pooling to consolidate visual tokens. 
TextMonkey~\cite{DBLP:journals/corr/textmonkey} performs compression based on token similarity. HiRED~\cite{arif2024hired} utilizes the attention map from the CLS token of ViT to discard unimportant tokens. These approaches rely on handcrafted metrics to gauge the importance of visual tokens, which does not guarantee optimal performance in terms of model accuracy. 
In contrast, our proposed method adopts a learning-based approach that allows the model to select the most appropriate visual scale for each local region. The selection is directly linked to the model's global optimization objective.
The third category~\cite{DBLP:journals/corr/VTW,DBLP:journals/corr/FastV}, directly applies token pruning methods from LLMs to the selection of visual tokens. These methods do not effectively utilize the inherent information. 
Moreover, the primary focus of these works is on enhancing computational efficiency, with less emphasis on the performance.

\section{Method}
The proposed \textsl{FocusLLaVA} is designed with two objectives: 
(i) fully leverage visual and textual information to effectively reduce the number of visual tokens for efficiency improvement; 
(ii) enable the model to learn to remove redundant visual information in a coarse-to-fine manner for performance improvement.

% \subsection{Preliminary}
% The concurrent MLLMs generally employ similar model structures, consisting of a visual image encoder, a connector, and a LLM. 
% The visual encoder typically uses the Vision Transformer (ViT) architecture and is initialized with the visual encoder from CLIP. 
% Various options exist for the connector, including Q-former~\cite{DBLP:conf/icml/BLIP2}, cross-attention, and MLP. 
% MLLMs like LLaVA~\cite{lan2024avgllava} have demonstrated that employing a simple MLP as the connector can also achieve good performance. 
% For a given image and its corresponding text query, the visual encoder first encodes the image into visual tokens. 
% These visual tokens are aligned to the LLM's word embedding space through the connector. 
% At the same time, the text corresponding to the image is processed through a tokenizer to obtain text embeddings. 
% The visual embeddings and text embeddings are then concatenated to form a unified sequence. 
% Finally, this unified sequence is input into the LLM's decoder for processing, outputting the corresponding answer.

\subsection{Overall Structure}
% In current research on various MLLMs, enhancing the resolution of visual images to improve model performance on fine-grained tasks has become a common practice. 
% However, high-resolution image inputs often significantly increase the number of visual tokens fed into LLMs, not only leading to increased computational costs and memory consumption but also amplifying irrelevant redundant information, which can adversely affect the model's performance.

% Our objective is to design a method, based on existing mainstream architectures of MLLMs, that can dynamically select visual tokens according to both visual and textual information. 
% This approach aims to maintain high resolution to handle complex tasks while filtering out the interference from redundant information through visual token selection, thereby saving computational resources. 
The overall structure of the proposed model is illustrated in \cref{fig:main}. We build our model based on the LLaVA-Next~\cite{Open-LLaVA-NeXT} with two core modules, vision-guided sampler and text-guided sampler inserted in the projector and LLM respectively.
The vision-guided sampler selects visual tokens based on the image information itself, while the latter combines textual information to provide stronger semantic guidance, filtering visual tokens relevant to user instructions. 

For a given image $\mathbf{I} \in \mathbb{R}^{H \times W \times 3}$ with high resolution, we follow LLaVA-Next to first segment the image into several local sub-images.
The segmented sub-images, along with the original image, are resized to a uniform size to form an image sequence $[\mathbf{I}_g, \mathbf{I}_0,  \mathbf{I}_1, \mathbf{I}_2, \ldots, \mathbf{I}_{N-1}]$, where $\mathbf{I}_g$ is the resized image block of the original image and $N$ is the total number of local sub-images. 
Each image in the sequence is encoded by ViT independently. The results are then passed through a projector, resulting in a set of image embeddings that are aligned with LLM's embedding space. 
Each image embedding is also called a visual token.
Then, we concatenate and reshape all visual tokens from all sub-images to form a global feature map, referred as $\mathbf{X} \in \mathbb{R}^{H_x \times W_x \times C}$. 
Subsequently, the vision-guided sampler performs visual token reduction for the feature map, with the remaining tokens inputted into the LLM. 
Within the LLM, a second round of selection is performed by the text-guided sampler, resulting in a set of visual tokens that are precisely correlated with text instruction.

% We partition the feature map by region to form a set of local feature blocks ${\mathbf{X}_0, \mathbf{X}_1, ..., \mathbf{X}_{M-1}}$, where $M$ is the number of local feature blocks. We denote the region's window size as $w$.  Each feature block $\mathbf{X}_i \in \mathbb{R}^{w \times w \times C}$ serves as the basic unit for visual token selection.

\subsection{Vision-Guided Sampler}
\label{sec:vgs}
For the global feature map $\mathbf{X}$, it is first partitioned by region to form a set of local feature blocks ${\mathbf{X}_0, \mathbf{X}_1, ..., \mathbf{X}_{M-1}}$, where $M$ is the number of local feature blocks. We denote the region's window size as $w$. Each feature block $\mathbf{X}_i \in \mathbb{R}^{w \times w \times C}$ serves as the basic unit for vision-guided sampler.
Note that the selection of visual scales is performed for each local region rather than the entire feature map. 
This design allows for adaptive scale selection for each local area of high-resolution images.
The vision-guided sampler processes feature block in two steps. First, it down-samples each feature block to multiple different scales. 
Then, it dynamically selects one of the visual scales based on the local information and global information.

\noindent\textbf{Multi-scale down-sampling.}
For each of the feature block $\mathbf{X}_i \in \mathbb{R}^{w \times w \times C}$, we use a set of Max-Pooling operations to down-sample it into several different scales. 
In our work, we use a window size of 4 to partition the feature map, resulting in $\mathbf{X}_i \in \mathbb{R}^{4 \times 4 \times C}$. We use three max-pooling operations, each with size $4 \times 4$, $2 \times 2$, and $1 \times 1$ to process the feature block, resulting in three kinds of token set for this feature block: $\mathrm{DS}[0](\mathbf{X}_i) \in \mathbb{R}^{1 \times 1 \times C}$, $\mathrm{DS}[1](\mathbf{X}_i) \in \mathbb{R}^{2 \times 2 \times C}$, and $\mathrm{DS}[2](\mathbf{X}_i) \in \mathbb{R}^{4 \times 4 \times C}$, where $\mathrm{DS}[i]$ denotes the $i$-th down-sampling type.
The design of max-pooling operations can be more flexible. For example, we can use unsymmetrical pooling with ratio 2 or $\frac{1}{2}$ to get more kinds of token set. We will discuss this design in the experiments.

\noindent\textbf{Multi-scale selection.}
We adopt the design concept of Mixture of Experts (MoE), treating multiple scales of down-sampling as experts. 
The goal is to select one expert for each local feature block.
To achieve this, we design a multi-scale selector by modeling the correlation between each local feature block and the global context.
Specifically, we use the scaled original image as $\mathbf{I}_g$ and its encoded features as $\mathbf{X}_g$. We flatten $\mathbf{X}_g$ to make $\mathbf{X}_g \in \mathbb{R}^{H_x W_x \times C}$.
Then, each basic feature block $X_i$ is pooled to $1 \times 1 \times C$, which is used to calculate the inner product with $\mathbf{X}_g$ as $\mathbf{Score}_i=pool(\mathbf{X}_i)^T\mathbf{X}_g$, where $\mathbf{Score}_i \in \mathbb{R}^{1 \times H_x W_x}$. Subsequently, a fully connected layer is applied to $\mathbf{Score}_i$ to predict the scale of each feature block, producing logits $\mathbf{Z} \in \mathbb{R}^{1 \times S}$ for visual scale selection. 
Finally, we apply the softmax function to calculate the probabilities of each visual scale. 
The tokens corresponding to the visual scale with the highest probability are then input to the LLM.

\noindent\textbf{Objective.}
Due to the non-differentiable nature of the selection operation in the down-sampling paths, the parameters of the selector cannot be effectively trained. To fix this, we introduce some techniques.
Given a local base feature block $\mathbf{X}_r$, 
we use $\mathrm{DS}[i]$ to denote the $i$-th down-sampling path. The final visual token corresponding to $\mathbf{X}_r$ that is ultimately selected and input into the LLM is 
\begin{align}
    \tilde{\mathbf{X}}_r = \mathrm{DS}[\mathrm{argmax}(\mathbf{Z})](\mathbf{X}_r),
\end{align}
where $\mathbf{Z}$ is the predicted logits corresponding to $\mathbf{X}_r$.
During inference, we apply the multi-scale selection using the aforementioned formula. While during training, we multiply the calculated probability with the tokens to make the entire process differentiable. Specifically:
\begin{align}
    \tilde{\mathbf{X}}_r = \mathrm{Top1}(\mathrm{Softmax}(\mathbf{Z})) * \mathrm{DS}[\mathrm{argmax}(\mathbf{Z})](X_r).
\end{align}

In this way, the probability calculated in the sidetrack is incorporated into the token, allowing the parameters of the selector to continue receiving gradient feedback through the LLM's optimization loss.
Furthermore, we observed that the selector tends to easily degrade into a trivial state where it continuously selects one particular branch during the training process. 
To counteract the laziness of network optimization, we follow switch-transformer~\cite{fedus2022switch} to introduce a balance loss to force the network to choose different branches. 
The form of the balance loss is as follows:
\begin{align}
&L_{balance} = \alpha \sum_{0}^{n} f_i * P_i,\\
&f_i = \frac{1}{N}\sum_{0}^{N-1}\textbf{1}(\mathrm{argmax}(\mathbf{Z})= i),\\
&P_i = \frac{1}{N}\sum_{\mathbf{Z}}\mathrm{Softmax}(\mathbf{Z})_i,
\end{align}
where $\alpha$ is the weight of the balance loss and $textbf{1}$ denotes the indicator function and $N$ denotes the total number of basic feature blocks within one batch.

% \noindent\textbf{Rethinking the Global Context}. 
% % We observe that in the global context, in addition to using the image itself, it is possible to integrate textual information. 
% In vision-guided sampler, we use the global image's features as the global context. Indeed, using textual information for visual selection is indeed an intuitive idea.
% For instance, SliME~\cite{zhang2024beyond} utilizes text embeddings to guide the selection of visual tokens. 
% In our initial design, we also attempted to use textual information as the global context in our Multi-Scale selection. 
% However, experiments revealed that its effectiveness was almost identical to using just image information as the global context. 
% We believe this is mainly because the text embeddings used in the global context do not provide enough visual semantic differentiation. 
% Understanding the intent of textual instructions and making judgments based on visual semantic information usually require alignment with visual tokens and complex reasoning. 
% Such differentiated text embeddings are challenging to obtain before being processed by a LLM. 
% Both LLM's word embeddings and text embeddings from other text models struggle to meet this requirement. 
% Based on these considerations, we discard the design of using textual information as guidance in this sampler.
% Instead, the guidance of text will be designed separately in subsequent sections.

\subsection{Text-Guided Sampler}
In text-guided sampler, we utilize the text comprehension capabilities of LLM to serve as the basis for visual token selection. 
To achieve this, text embeddings from the intermediate layers of LLMs are selected as guidance for visual token filtering.
We believe that the text embeddings in the intermediate layers of LLM, which have already interacted with the visual signal, contain sufficient semantic information and have the ability to select important visual tokens.
This idea is supported by a further analysis. As shown in \cref{fig:evolution}, we visualize the attention scores received by each visual token in several layers. 
We observed that in the shallow layers of the LLM, the importance pattern of visual tokens is unstable and undergoes significant changes. Moreover, the importance map can hardly focus on the most correlated area.
In contrast, in the intermediate and deeper layers of the LLM, the importance pattern of visual tokens tends to become consistent and accurate. 
For this reason, we argue that the text-guided sampler should be placed the middle layers of LLM, rather than shallow layers or out of LLM. 

For the i-th layer of the LLM, we first compute the similarity matrix $\mathbf{A}_i \in \mathbb{R}^{h \times N \times T}$ between the text prompt's embedding \( \mathbf{h}^t_i \) and the visual token's embedding \( \mathbf{h}^v_i \), where \( h \) denotes the number of heads in the multi-head attention, \( N \) indicates the number of visual tokens, and \( T \) represents the number of text prompt tokens:
\begin{align}
    &\mathbf{Q}_i = \mathbf{W}^\mathbf{q}_i\mathbf{h}^t_i \quad \mathbf{K}_i = \mathbf{W}^\mathbf{k}_i\mathbf{h}^v_i\\
    &\mathbf{A}_i = \mathrm{Softmax}\left(\frac{\mathbf{Q}_i\mathbf{K}_i^T}{\sqrt{d}}\right).
\end{align}

To determine the importance score \(\mathbf{S}_i\) for each visual token, we first employ a reduce-max operation to identify the highest score across all attention heads. 
Subsequently, we average these importance scores over all textual tokens. 
After obtaining the importance score  \(\mathbf{S}_i\), we select the top-k visual tokens with the highest importance scores for retention.
Mathematically, this process can be expressed as:
\begin{equation}
    \label{eq:importance}
    \mathbf{S}_i = \frac{1}{T} \sum_{j=0}^{T-1} \max_{h} \mathbf{A}_i[h, j], \quad \mathbf{S}_i \in \mathbb{R}^{N},
\end{equation}
where \(\mathbf{A}_i\) represents the attention scores for the \(i\)-th visual token across all attention heads and textual tokens, \(T\) is the total number of textual tokens, and \(N\) is the total number of visual tokens. 
In the selection of significant visual tokens, we first sort \(\mathbf{S}_i\) in descending order to obtain \(\{s_0, s_1, ..., s_{N-1}\}\). 
Then, we select the visual tokens whose cumulative importance exceeds a given threshold,
\begin{equation}
    k = \min_j \sum_{i=0}^{j-1} \frac{s_i}{\sum_{i=0}^{N-1} s_i} > \gamma,
\end{equation}
where \(\gamma\) is the importance threshold.
Finally, the top \(k\) visual tokens, ranked by the importance, are selected.

In LLMs, filtering visual tokens based on importance typically leads to a decline in model performance. 
We believe that discarding some tokens in LLMs causes inconsistency between the training and testing. 
To compensate for the performance degradation, we introduce randomness during the training process to enhance the model's adaptability. 
Specifically, during training, for each image-text pair, we can randomly select a decoder layer and an importance threshold \(\gamma\), and then choose the visual tokens according to the importance formula. 
The model, after being trained with this randomness-enhanced approach, demonstrates stronger robustness in testing.

\begin{table*}[]
\centering
\caption{
\textbf{Comparison with existing MLLMs on popular benchmarks.}
VQA\textsuperscript{\textit{T}}: TextVQA~\cite{DBLP:conf/cvpr/TextVQA}; SQA: ScienceQA~\cite{DBLP:conf/nips/sqa}; LLaVA\textsuperscript{\textit{W}}: LLaVA-bench-in-the-wild; MME\textsuperscript{\textit{P,C}}: Perception and Cognition in MME~\cite{DBLP:journals/corr/mme};
MMB\textsuperscript{\textit{C}} denotes MMBench-CN~\cite{DBLP:journals/corr/mmb}.
}
\resizebox{\textwidth}{!}{
    \begin{tabular}{llllllllllll}
\toprule
Method & LLM & VQA\textsuperscript{\textit{T}} & SQA  & GQA  & POPE & MM-Vet & LLaVA\textsuperscript{\textit{W}} & MME\textsuperscript{\textit{P}}  & MME\textsuperscript{\textit{C}} & MMB & MMB\textsuperscript{\textit{C}} \\
\midrule
Instruct-BLIP~\cite{DBLP:conf/nips/InstructBLIP} & Vicuna-7B  & 50.1                                    & -    & 49.2 & -    & 26.2   & 60.9    & 1084   & 229   & 36.0  & 23.7  \\
Qwen-VL~\cite{qwen-vl} & Qwen-7B & 63.8 & - & 59.3 & - & - & - & 1487.6 & - & 60.6 & 7.4 \\
LLaVA-1.5~\cite{Liu_2024_CVPR_llava1_5} & Vicuna-7B  & 58.2 & - & 62.0 & 85.9 & 30.5   & 65.4 & 1510 & - & 64.3  & 58.3  \\
LLaVA-1.5 & Llama3-8B  & 58.9                                    & -    & 61.9 & 85.1 & 34.8   & 70.5    & 1544   & 328   & 72.9  & 67.7  \\
mPlugOwl3~\cite{ye2024mplug3} & Qwen-8B & 69.0 & -    & 65.0 & 88.2 & 40.1 & -       & -      & -  & 77.6  & 74.3  \\ \hline
Otter-HD~\cite{DBLP:journals/corr/Otterhd} & Fuyu-8B & -                                       & -    & -    & 86.0 & -      & -       & 1223   & 331   & 58.30 & -     \\
LLaVA-NeXT~\cite{liu2024llavanext}              & Vicuna-7B  & 64.9                                    & 70.1 & 64.2 & 86.5 & -      & -       & 1519   & 332   & 67.4  & 60.6  \\
Mini-Gemini-HD~\cite{DBLP:journals/corr/Mini-Gemini} & Vicuna-7B  & 68.4                                    & -    & -    & -    & 41.3   & -       & 1546   & 319   & 65.8  & -     \\ \hline
SliME~\cite{zhang2024beyond}                     & Llama3-8B  & 64.7                                    & 84.2 & 63.9 & -    & 37.4   & 73.9    & 1578   & 337   & 75.0  & 71.8  \\
LLaVA-Prumerge+~\cite{DBLP:journals/corr/LLaVA-PruMerge} & Vicuna-7B  & 57.1                                    & 68.3 & -    & 84.0 &        & -       & 1462   & -     & 64.9  & -     \\
Trim~\cite{song2024less} & Vicuna-7B  & -                                       & 69.1 & 61.4 & 85.3 & 28.0   & 58.7    & 1461   & -     & 67.4  & 54.9  \\
HiRED~\cite{arif2024hired}                       & Vicuna-13B & 65.2                                    & 73.2 & -    & 87.7 & -      & -       & 1570   & -     & -     & -     \\
\midrule
LLaVA-NeXT (Ours) & Llama3-8B  & 69.4 & 77.3 & 65.7 & 86.9 & 40.6   & 64.7    & 1558   & 334   & 74.2  & 70.1  \\
\textsl{FocusLLaVA} (Ours) & Llama3-8B  & 70.0                                    & 79.0 & 66.0 & 87.7 & 41.3   & 65.6    & 1600   & 328   & 74.7  & 70.3  \\
\bottomrule
\end{tabular}
\label{tab:main_results}
}
\end{table*}

\section{Experiments}
\subsection{Implementation Details}
We use CLIP~\cite{DBLP:conf/icml/CLIP} ViT-L/14 as the visual encoder (default resolution 336 × 336), LLama3-8B as the LLM and a 2-layer MLP as the connector.
We keep the adaptive image slicing technique used in LLaVA-Next~\cite{Open-LLaVA-NeXT} with each sub-image being resized to $336 \times 336$.
For the vision-guided sampler, we set the window size of local feature block to 4. Three scales $4 \times 4$, $2 \times 2$ and $1 \times 1$ are used for multi-scale selection.
For the text-guided sampler, we insert it into the 8-th layer in LLM. $\gamma$ is set to 0.85 by default. During training, it is randomly inserted into layers between 8 and 24. 
Following LLaVA-Next, we use similar training settings for visual instruction fine-tuning. The learning rate is set to 1e-5 with a cosine learning rate scheduler and a batch size of 128. The training takes one epoch using AdamW optimizer. Deepspeed Zero2 is used as the training framework. 
We conduct experiments on a server equipped with 8 Nvidia A100 GPUs, each with 80GB of VRAM.

\subsection{Datasets and Benchmarks}

\noindent\textbf{Training datasets.}
We pretrain and fine-tune the proposed model with only open-source data.
For pretraining, we follow the popular llava series to use llava-pretrain-558k.
For fine-tuning, we have made every effort to align our training data with LLaVA-NeXT~\cite{liu2024llavanext}. However, as the tens of thousands of real user data used by LLaVA-NeXT is not released, we follow Open-LLaVA-NeXT~\cite{chen2024open} to use 200K ALLaVA-Instruct-VFLAN-4V~\cite{chen2024allava} data as a substitute. Additionally, since TextVQA~\cite{DBLP:conf/cvpr/TextVQA} has been included in the training data of most existing LMMs, we retain it to enable fair comparisons with other LMMs.
As a result, the fine-tuning data includes 1M samples, covering sharegpt4v-mix665k~\cite{chen2023sharegpt4v},  ALLaVA-Instruct-VFLAN-4V, DocVQA~\cite{DBLP:conf/wacv/docvqa}, SynDog-EN~\cite{DBLP:conf/eccv/SynDog-EN}, ChartQA~\cite{DBLP:conf/acl/chartqa}, DVQA~\cite{kafle2018dvqa}, AI2D~\cite{DBLP:conf/eccv/ai2d}, and GeoQA+~\cite{chen2021geoqa}.
The specific data configuration are available in the supplementary.

\noindent\textbf{Benchmarks.} We use 10 popular benchmarks to evaluate our method, including (1) General question answering benchmarks such as GQA~\cite{DBLP:conf/cvpr/GQA} and ScienceQA~\cite{DBLP:conf/nips/sqa}; (2) Optical character based visual question answering benchmark such as TextVQA~\cite{DBLP:conf/cvpr/TextVQA};
(3) MLLM benchmarks for specific abilities, like POPE~\cite{DBLP:conf/emnlp/POPE}, MM-Vet~\cite{yu2024mm-vet} and LLaVA-in-the-wild~\cite{DBLP:conf/nips/LiuLWL23a_llava};
(4) Comprehensive MLLM benchmarks such as MME~\cite{DBLP:journals/corr/mme} (Perception and Cognition), MMBench~\cite{DBLP:journals/corr/mmb} and MMBench-CN.

\subsection{Performance}
We have selected three types of approaches for comparative reference, with their results presented in \cref{tab:main_results}, separated by different categories. The first type consists of various mainstream MLLMs. The second type includes MLLMs that take high-resolution inputs. The third type focuses on works dedicated to compressing visual tokens, for which we have chosen their best outcomes. The final section presents our baseline and methods. Given that the foundational LLMs, training datasets, and training configurations used in current MLLM research vary, most of the results in \cref{tab:main_results} are not directly comparable. 
However, these results can serve as a reference to illustrate the approximate performance level of our implemented baseline and methods. In the third part, specifically for our self-implemented LLaVA-Next and \textsl{FocusLLaVA}, we ensured strict alignment in the foundational large models, training datasets, and training settings, making them fairly comparable. 
Our findings indicate that our implemented baseline is highly competitive compared to both mainstream and high-resolution MLLMs. Furthermore, our proposed \textsl{FocusLLaVA} demonstrates clear improvements over this baseline.
% \noindent\textbf{Baselines}. We compare our method with several types of baselines. (1) General Baselines. We select Qwen-VL~\cite{qwen-vl}, LLaVA1.5~\cite{Liu_2024_CVPR_llava1_5},LLaVANext~\cite{liu2024llavanext}, Mini-GPT-v2, Shikra, BLIP2~\cite{DBLP:conf/icml/BLIP2}, and instructBLIP~\cite{DBLP:conf/nips/InstructBLIP} as representative general baselines. (2) High-resolution MLLMs. We select HiRED~\cite{arif2024hired}, SliME~\cite{zhang2024beyond}, LLaVA-UHD~\cite{xu2024llava-uhd} to compare. These methods are proposed for visual token compression. (3) The re-implemented LLaVANext, which serves as the direct baseline as it is trained under the identical setting with our method.

% \noindent\textbf{Performance on Standard Benchmarks}.

% \noindent\textbf{Trade-off between Performance and Efficiency}.

\begin{figure}[ht]
    \centering
    \includegraphics[width=0.95\linewidth]{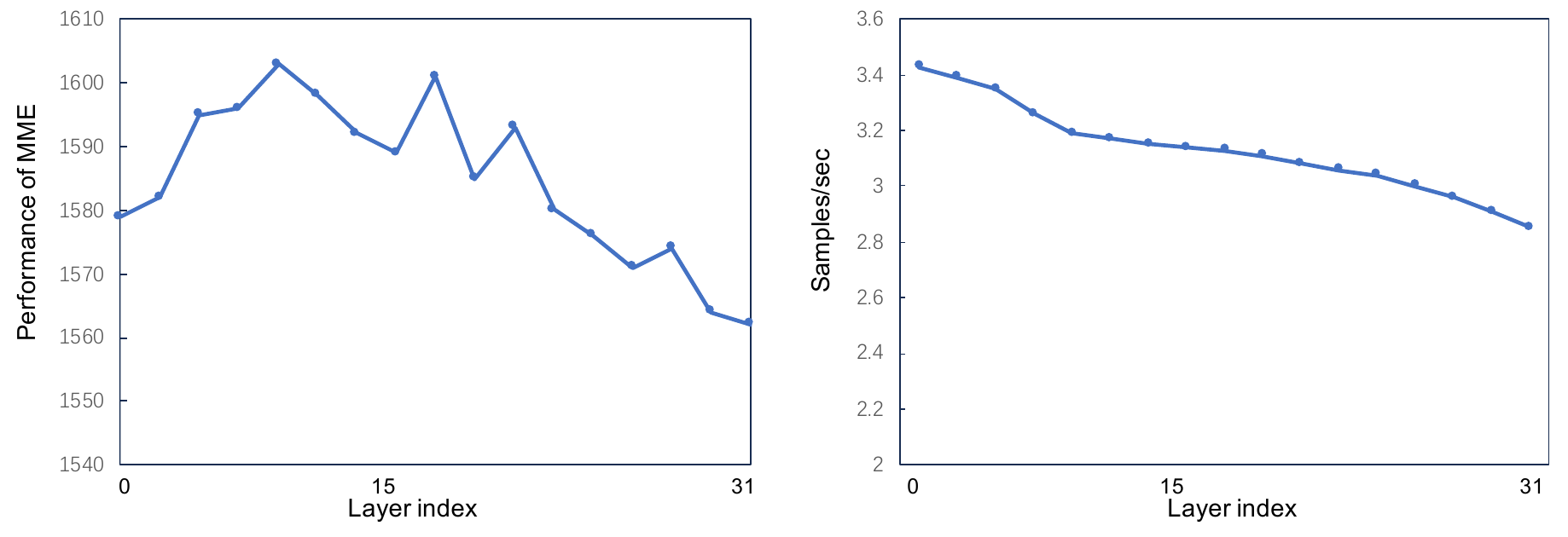}
    \caption{\textbf{Performance and speed with textual guidance in different layers.}}
    \label{fig:layer}
\end{figure}

\subsection{Analysis}
For experiments in this part, a smaller training set (200K) is used for convenience. It is randomly sampled from the default training set.

\noindent\textbf{Ablation on visual and textual guidance.}
We experiment to explore the individual contribution of vision-guided sampler and text-guided sampler.
The results are presented in \cref{tab:ablation}. The findings indicate that both the vision-guided sampler and the text-guided sampler help the model improve performance across multiple evaluation benchmarks. When combined, they further enhance the performance on these benchmarks. Besides, the efficiency is further improved. Note that since the visual tokens removed by the text-guided sampler still go through several layers of computation, the number of remaining tokens is calculated through conversion. Suppose there are $m$ visual tokens and text-guided sampler removes $n$ tokens in $i$-th layer, then the number of remaining tokens is calculated through $m - n + i*n/32$, where 32 is the total number of layers.
\begin{table}[]
    \centering
    \caption{
    \textbf{Ablation on visual and textual guidance.}
    ``Vision tokens'' means the percentage of visual tokens remained. ``Speed'' means number of samplers per second. The speed is tested on TextVQA~\cite{DBLP:conf/cvpr/TextVQA}.}
    \resizebox{\linewidth}{!}{
        \begin{tabular}{lcccccc}
            \toprule
            Setting  & \makecell{Vision\\tokens} & Speed & TextVQA       & MME           & ScienceQA     & GQA           \\ \midrule
            Baseline & 100\%    & 2.85  & 65.5          & 1562          & 76.4          & 61.7          \\
            Vision   & 49\%     & 3.76  & 65.2          & 1590          & 77.8          & 62.3          \\
            Text     & 81\%     & 3.19  & 65.4          & \textbf{1603} & 76.9          & 62.1          \\
            Both     & 39\%     & 4.01  & \textbf{65.9} & 1597          & \textbf{78.1} & \textbf{62.6} \\ \bottomrule
        \end{tabular}
    }
    \label{tab:ablation}
\end{table}

\noindent\textbf{More scales.}
In this part, we investigate the effect of additional scales. Building upon the original three scales, 4x4, 2x2, and 1x1, we add four asymmetric scales: 4x2, 2x4, 2x1, and 1x2. The results are presented in \cref{tab:more_scales}. Here, ``Baseline'' refers to our reproduction of the LLaVA-Next, ``3-Branch'' represents the default implementation of our method, and ``7-Branch'' indicates the extension to 7 visual scales. The results show that adding more scales leads to a notable 1.1\% increase in TextVQA over the ``3-branch''. We attribute this enhancement to the fact that the TextVQA task requires the recognition of fine-grained information, such as text, and the inclusion of more scales aids the model in better handling these detailed elements.

\begin{table}[]
    \centering
    \caption{
    \textbf{More flexible pooling designs.}
    * denotes default.}
    \resizebox{0.75\linewidth}{!}{
        \begin{tabular}{lcccc}
            \toprule
            Setting                             & TextVQA       & MME           & ScienceQA     & GQA           \\ \midrule
            Baseline                            & 65.5          & 1562          & 76.4          & 61.7          \\
            3-branch*                            & 65.2          & \textbf{1590} & \textbf{77.8} & 62.3          \\
            7-branch                            & \textbf{66.6} & 1589          & 77.2          & \textbf{62.4}          \\ \bottomrule
        \end{tabular}
    }
    \label{tab:more_scales}
\end{table}

\noindent\textbf{Window size of feature block}
In this part, we compare the impact of different window size of feature block in the vision-guided sampler. In the default setting, the feature block size is 4x4. We add several sizes for comparison, including 1x1, 2x2, 8x8, and 12x12. For the 1x1 feature block, since it is no longer possible to perform multi-scale down-sampling, we directly predict whether it should be retained or discarded. For the 2x2 feature block, only 1x1 and 2x2 scales can be used for down-sampling. For 8x8 and 12x12, the down-sampling branches used are consistent with the default configuration. The experimental results are shown in \cref{tab:window_size}. The results indicate that setting the window size too small  or too large is harmfull to performance. Window size such as 1x1 and 2x2, even leads to a decrease in performance compared to the baseline. Overall, selecting 4x4 as the feature block size yields the best performance. We believe that as the feature map size decreases, the regions are divided more finely, but the number of available multi-scale options also decreases. Conversely, as the size increases, the region division becomes coarser, and when the scale reaches the same size as the feature map, all regions are down-sampled in the same manner.

\begin{table}[]
    \centering
    \caption{
    \textbf{Influence of size of the local feature block.}
    * denotes default.}
    \resizebox{0.8\linewidth}{!}{
        \begin{tabular}{lcccc}
            \toprule
            Setting      & TextVQA       & MME           & ScienceQA     & GQA           \\ \midrule
            baseline     & \textbf{65.5} & 1562          & 76.4          & 61.7          \\
            1x1          & 64.5          & 1557          & 73.9          & 61.9          \\
            2x2          & 65.1          & 1549          & 76.1          & 61.7          \\
            4x4*        & 65.2          & \textbf{1590} & 77.8          & \textbf{62.3} \\
            8x8          & 64.9          & 1570          & 77.2          & 61.8          \\
            12x12       & 64.2          & 1567          & \textbf{78.2} & 61.5          \\ \bottomrule
        \end{tabular}
    }
    \label{tab:window_size}
\end{table}

\begin{figure}[ht]
    \centering
    \includegraphics[width=0.95\linewidth]{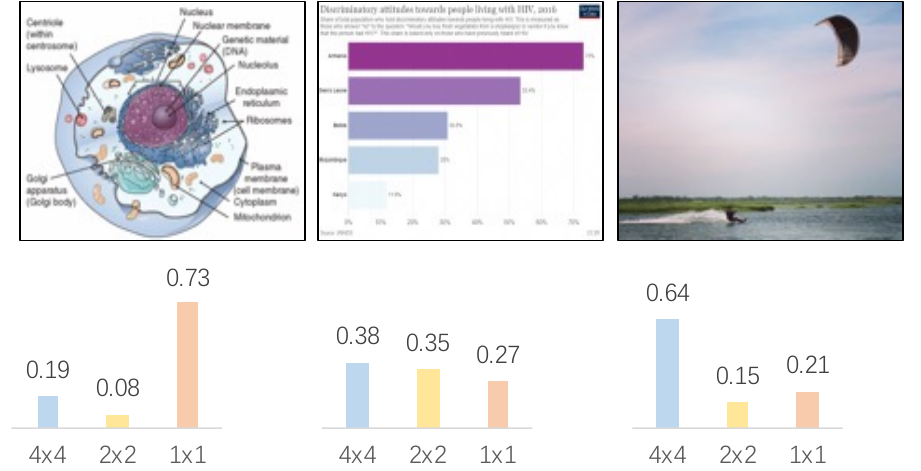} % 修改为你自己的图片文件路径
    \caption{Statistics of multi-scale sampling. For each sample, the proportion of the three types of max-pooling is shown.}
    \label{fig:ratio}
\end{figure}

\noindent\textbf{Role of balance loss.} 
In this part, we investigate the role of the balance loss in multi-scale selection. 
The original balance loss is designed to ensure that each expert is selected uniformly in MoE structure. 
In our scenario, the balance loss encourages the model to explore different visual scale selections, thereby training the sampler module.
It is important to note that after being trained with the balance loss, the model does not uniformly select each scale. It adjusts according to the actual situation. For example, as shown in \cref{fig:ratio}, the proportion of different visual scales chosen by the model varies. In the leftmost image, which contains a lot of text detail, a higher proportion is allocated to 1x1 pooling, meaning no down-sampling is performed. Conversely, in the rightmost image, which has large background areas, a greater proportion is assigned to 4x4 pooling. To further analyze the effect of the balance loss, we conducted an ablation study, and the results are presented in \cref{tab:balance}.
Based on the results, we find that the setting without balance loss get poor results. We dig deep and find that the selector degrades to consistently select a single scale. For example, if the selector is initialized to select one scale, then this choice won't be changed during training.

\begin{table}[]
    \centering
    \caption{
    \textbf{Role of load balance loss.}
    * denotes default.}
    \resizebox{0.9\linewidth}{!}{
        \begin{tabular}{lcccc}
        \toprule
        Setting             & TextVQA & MME  & ScienceQA & GQA  \\ \midrule
        w/o                 & 61.9    & 1439 & 74.6      & 57.1 \\
        weight 0.01         & 63.7    & 1537 & 75.9      & 60.0 \\
        weight 0.1*    & 65.2    & 1590 & 77.8      & 62.3 \\
        weight 0.5          & 65.0    & 1576 & 76.9      & 61.5 \\ \bottomrule
        \end{tabular}
    }
    \label{tab:balance}
\end{table}

\noindent\textbf{Effect of textual guidance.}
In this part, we explore the effect of textual guidance. We set up four settings in our experiments. (1) Baseline: train using the original structure of LLaVA-Next. (2) Direct: Based on the baseline, use the text-guided sampler without training. \(\gamma\) is set to 90\%. (3) Train: Perform the full Instruction-Tuning with text-guided sampler. (4) Random: Based on ``Train'', add augmentation to the layers and \(\gamma\) values.
The results are presented in \cref{tab:text_guidance}. We observe that using textual guidance without training leads to a noticeable performance drop on the TextVQA, MME, and GQA. After training, the introduction of the Text-guided Filter hardly causes any performance reduction, and even shows slight improvements on MME, ScienceQA, and GQA. 
With the introduction of the random training technique, we observe clear improvements over the Baseline on MME and GQA. These experimental results indicate that incorporating the training process significantly mitigates the performance drop caused by the text-guided sampler, and our introduced random training technique further enhances the robustness of the process.

\begin{table}[]
    \centering
    \caption{
    \textbf{Impact of textual guidance.}
    * denotes default.}
    \resizebox{0.8\linewidth}{!}{
        \begin{tabular}{lcccc}
            \toprule
            Setting                             & TextVQA       & MME           & ScienceQA     & GQA           \\ \midrule
            Baseline                            & \textbf{65.5}          & 1562          & 76.4          & 61.7          \\
            Direct                              & 64.9          & 1543          & 76.6          & 60.8          \\
            Train                               & 65.3          & 1586          & 77.3          & 61.9          \\
            Random*                          & 65.4 & \textbf{1603} & \textbf{76.9} & \textbf{62.1} \\ \bottomrule
        \end{tabular}
    }
    \label{tab:text_guidance}
\end{table}

\begin{figure*}[!ht]
    \centering
    \includegraphics[width=0.9\textwidth]{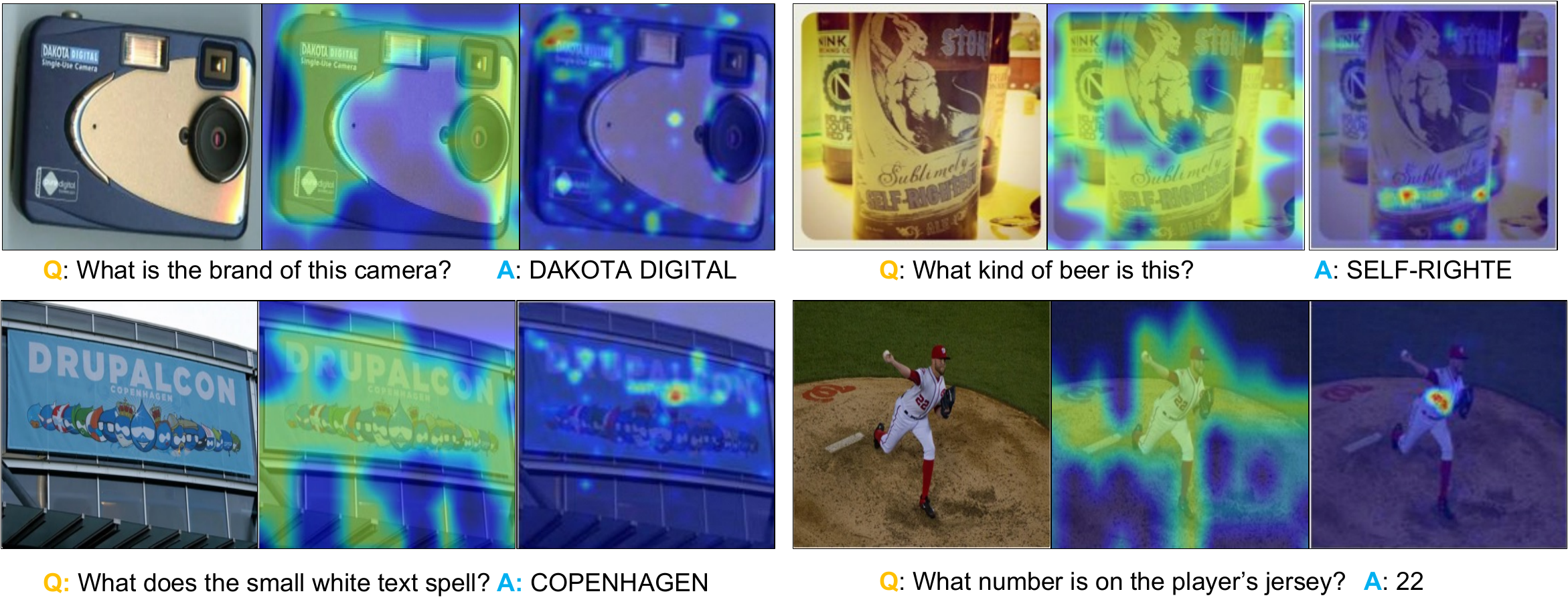} % 修改为你自己的图片文件路径
    \caption{
    \textbf{Heatmap of selected areas from vision-guided sampler and text-guided sampler.}
    For each image set, from left to right: the original image, the heatmap of vision-guided sampler, the heatmap of text-guided sampler.}
    \label{fig:select_region} % 你可以根据需要添加标签
\end{figure*}

\noindent\textbf{Textual guidance in different layers.}
We investigated the impact of placing a text-guided sampler at different layers within a LLM. The results are presented in Figure-\ref{fig:layer}. 
Notably, positioning the sampler at the layer 31, which is the final layer of the LLM, is equivalent to not performing any reduction operation. Our findings reveal that as the layer number increases, the model's inference speed gradually decreases. The model's performance initially improves before eventually decline. This suggests that the text-guided sampler performs better when placed in the middle layers, aligning with our initial expectations.

\noindent\textbf{What has been selected?} 
In this part, we discuss the roles of the vision-guided sampler and the text-guided sampler in the selection of visual tokens. 
We have selected several cases to observe the critical regions chosen by each module, with the results illustrated in \cref{fig:select_region}. 
For vision-guided sampler, we also select some examples to compare their proportion of each scale. It is shown in \cref{fig:ratio}.
The results reveal that the vision-guided sampler tends to focus on areas of high information density within the image. 
Specifically, it retains elements such as text, patterns, and people. 
In contrast, the text-guided sampler emphasizes regions directly related to the query. 
For instance, when asked about a mobile phone's brand, it highlights the textual area containing the answer. 
Similarly, when inquired about the number on a player's jersey, it disregards most other regions and precisely captures the area where the answer is located.
% Through observation, we believe that the combination of a vision-guided sampler followed by a text-guided sampler can form a two-stage, coarse-to-fine pipeline for selecting visual tokens.

\begin{figure}[ht]
    \centering
    \includegraphics[width=0.98\linewidth]{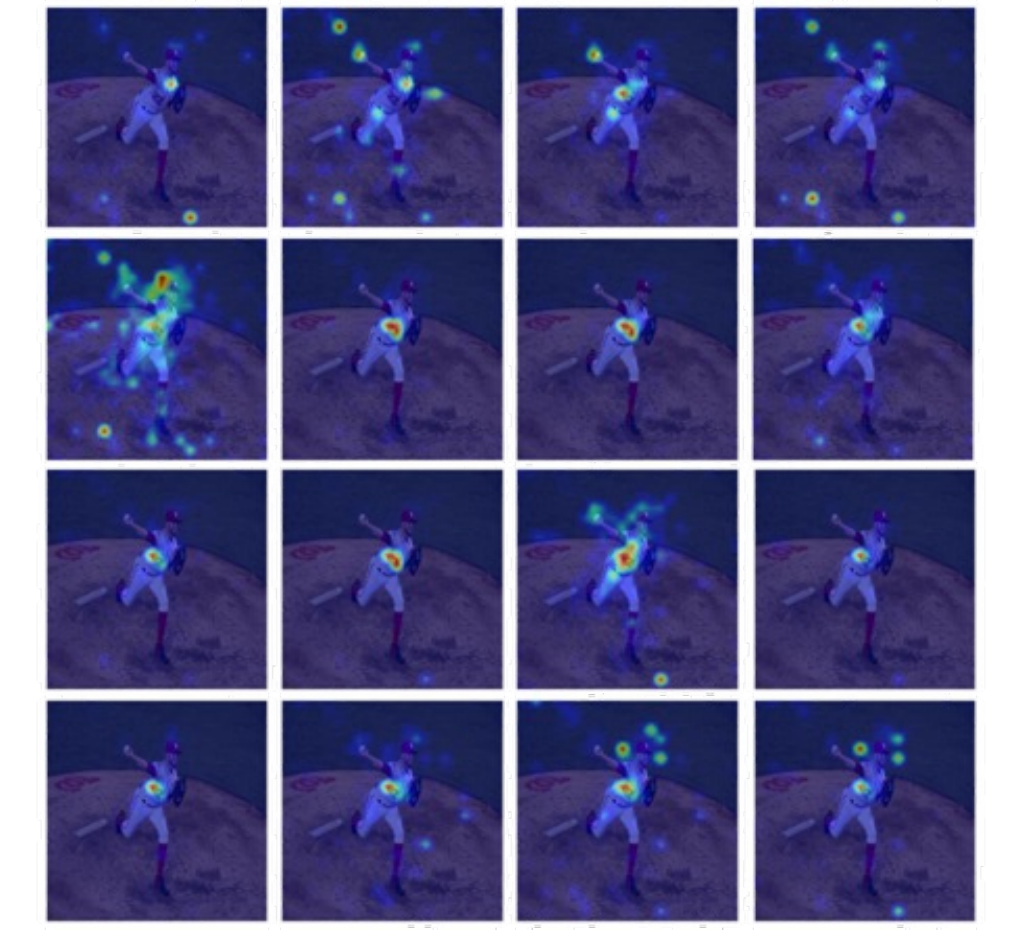} % 修改为你自己的图片文件路径
    \caption{
    \textbf{The importance map from different layers of LLM.}
    There are 32 layers in total. We select the importance map every two layers. The maps are arranged in reading order.}
    \label{fig:evolution} % 你可以根据需要添加标签
\end{figure}
\noindent\textbf{Evolution of textual guidance.}
We explore the differences of importance map defined by \cref{eq:importance} across various layers of the LLM, and the results are presented in \cref{fig:evolution}. 
We observed that in the shallow layers of the LLM, textual guidance fails to focus on the regions where the answers are located, and the areas of focus also change frequently. 
As the layer goes deeper, the regions focused on by textual guidance become increasingly accurate and gradually stabilize.
This indicates that the LLM's understanding of the association between image and text information is a progressive process. 
Performing filter operations too early will directly discard the image regions containing the answers, thereby affecting the model's performance, especially in scenarios requiring attention to fine-grained tasks. 
This also demonstrates the necessity of placing the text-guided sampler in the middle layers of the LLM.
On the other hand, since textual guidance in the shallow layers of the LLM is not enough for effectively performing token selection, using visual guidance to remove redundancy before entering LLM is a good complement. 
% It effectively eliminates some noise regions that do not need much attention.

\section{Conclusion}

In this work, we propose \textsl{FocusLLaVA}, which removes visual redundancy adaptively and simultaneously improves both performance and efficiency.
\textsl{FocusLLaVA} fully leverages both visual and textual information as guidance in its design, forming a coarse-to-fine pipeline for visual token compression. Specifically, it uses visual information to compress redundant regions with low information density and use the textual information to select visual tokens that are strongly correlated with the user instruction.
\textsl{FocusLLaVA} achieves improvement both in efficiency and performance.
Extensive experiments are conducted on various mainstream multimodal benchmarks, to validate the effectiveness of the proposed method.

{
    \small
    \bibliographystyle{ieeenat_fullname}
    \bibliography{main}
}


\maketitlesupplementary

\section{Comparison with Heuristic Visual Token Dropping}
In this part, we compare the proposed method with the manually designed metrics. This is used to illustrate that a learned metric rather than hand-crafted will solve the problem of performance reduction.
To this end, we modify the vision-guided sampler by removing its mechanism for predicting scales and instead select important tokens based on manually designed similarity measures. Specifically, for each feature block, we flatten it to obtain \(\mathbf{X}_i \in \mathbb{R}^{16 \times C}\), and then compute the similarity matrix with the global image feature block \(\mathbf{X}_g \in \mathbb{R}^{H_xW_x \times C}\) to get \(A_i = \mathbf{X}_g^T\mathbf{X}_i \in \mathbb{R}^{H_xW_x \times 16}\). We then calculate the importance score for each token in the feature block by averaging: \(Score_i = \frac{1}{H_xW_x} \sum_{j=1}^{H_xW_x} A_{ij}\). Finally, we concatenate all \(Score_i\) and select the top visual tokens based on these scores.
Our experimental results, as shown in \cref{tab:heuristic}, indicate that using manually designed methods leads to a decrease in model performance, regardless of the number of visual tokens retained. In contrast, FocusLLaVa achieves performance improvements even when reducing the number of visual tokens.
\begin{table}[h]
    \centering
    \caption{
    \textbf{Comparison with heuristic methods.}}
    \resizebox{0.9\linewidth}{!}{
        \begin{tabular}{lccccc}
            \toprule
            Setting    & \#Visual Tokens & TextVQA       & MME           & ScienceQA     & GQA           \\ \midrule
            Top-100\%  & 100\%           & \textbf{65.5} & 1562          & 76.4          & 61.7          \\
            Top-80\%   & 80\%            & 65.2          & 1571          & 76.2          & 61.3          \\
            Top-60\%   & 60\%            & 64.3          & 1549          & 75.9          & 58.9          \\
            Top-40\%   & 40\%            & 62.6          & 1508          & 75.2          & 58.4          \\
            FocusLLava & 49\%            & 65.2          & \textbf{1590} & \textbf{77.8} & \textbf{62.3} \\ \bottomrule
        \end{tabular}
    }
    \label{tab:heuristic}
\end{table}

\section{More Examples}
\begin{figure}[t]
    \centering
    \includegraphics[width=0.8\linewidth]{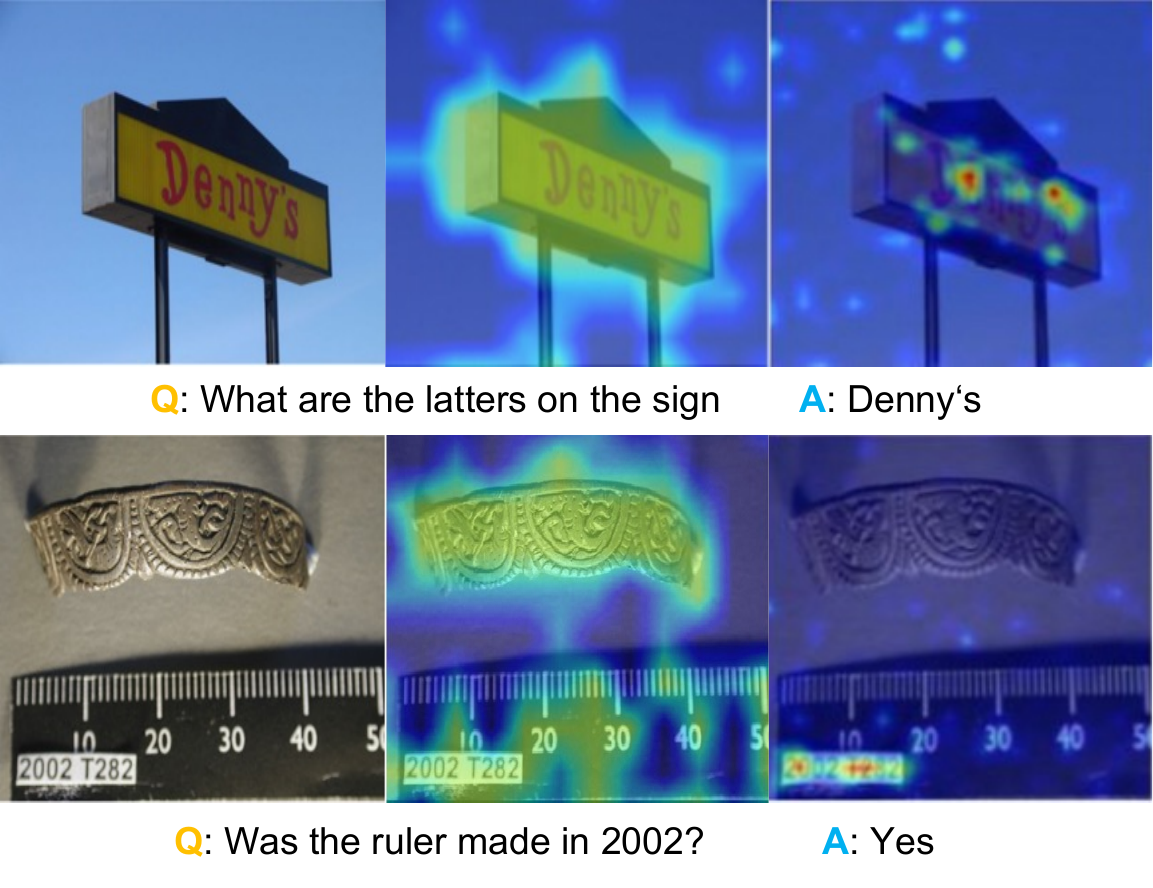}
    \caption{\textbf{Selected areas from vision-guided sampler and text-guided sampler.} }
    \label{fig:more_example1}
\end{figure}

\begin{figure}[t]
    \centering
    \includegraphics[width=0.8\linewidth]{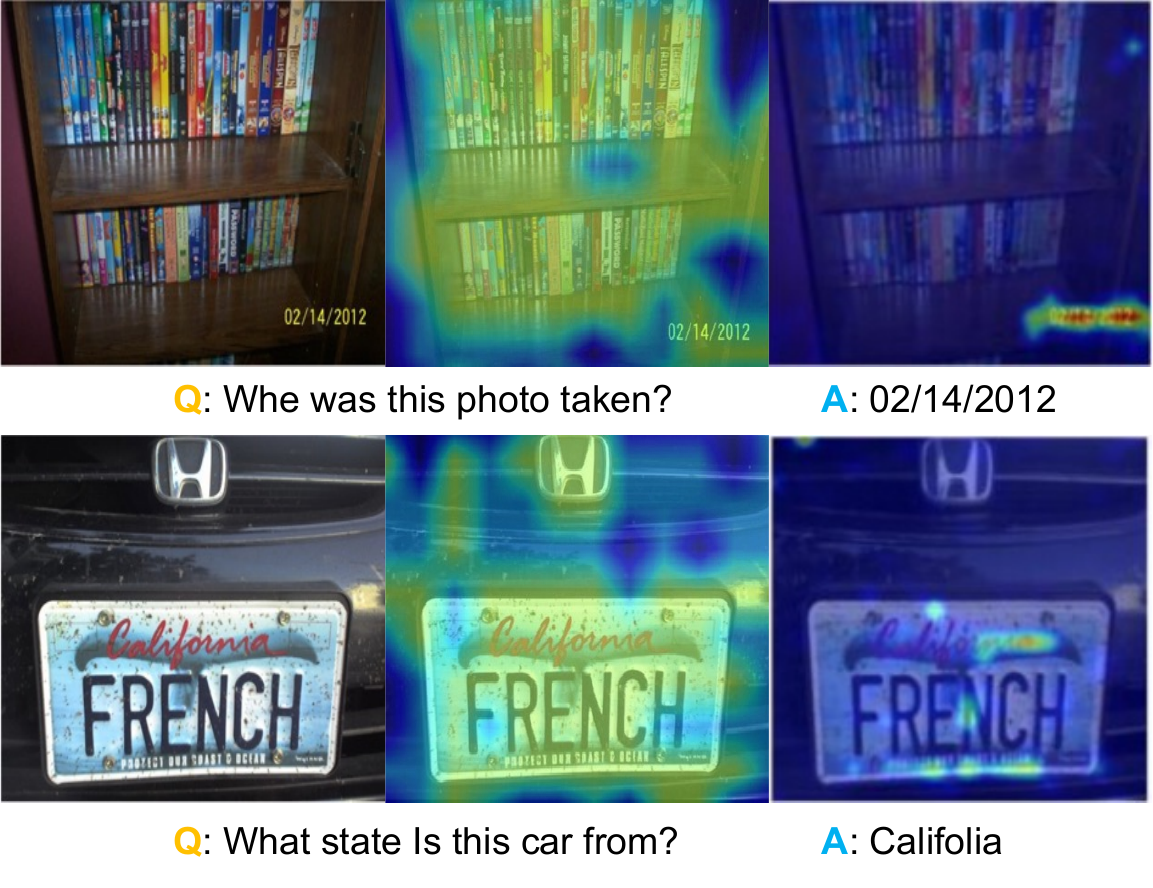}
    \caption{\textbf{Selected areas from vision-guided sampler and text-guided sampler.} }
    \label{fig:more_example2}
\end{figure}

\begin{figure}[h]
    \centering
    \includegraphics[width=0.8\linewidth]{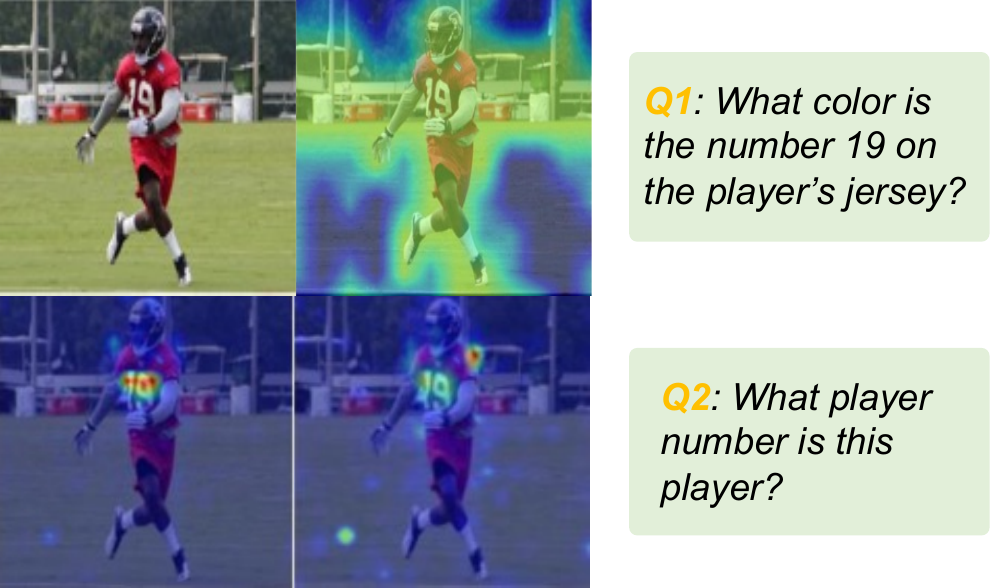}
    \caption{\textbf{Difference of selected areas across different queries.}}
    \label{fig:different_queries1}
\end{figure}

\begin{figure}[h]
    \centering
    \includegraphics[width=0.8\linewidth]{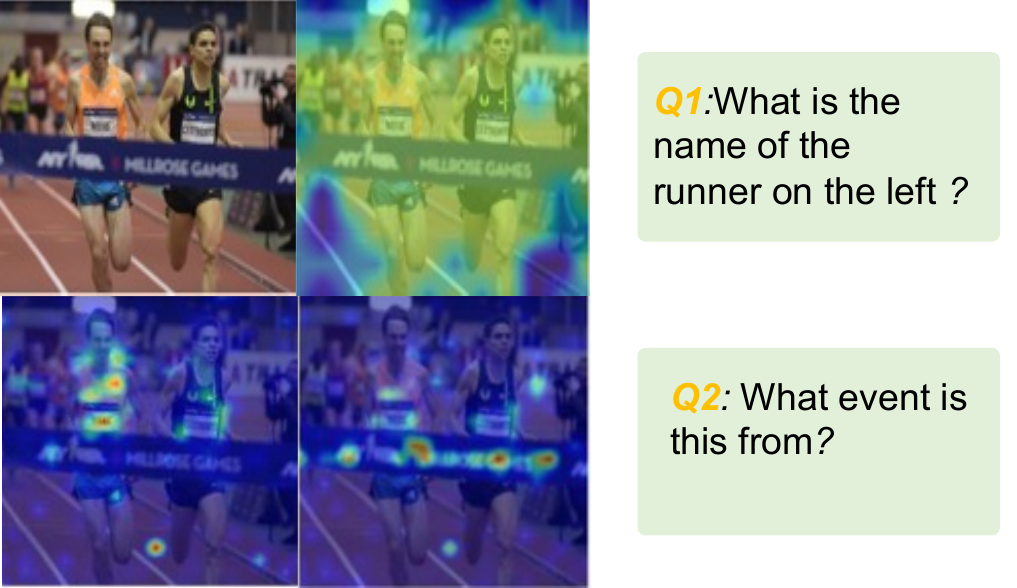}
    \caption{\textbf{Difference of selected areas across different queries.}}
    \label{fig:different_queries2}
\end{figure}

In this part, more examples are visualized in \cref{fig:more_example1} and \cref{fig:more_example2} to demonstrate the different characteristics of visual guidance and textual guidance. Besides, we also visualize the different areas selected across different instructions. It is shown in \cref{fig:different_queries1} and \cref{fig:different_queries2}. 

\begin{figure}[h]
    \centering
    \includegraphics[width=0.8\linewidth]{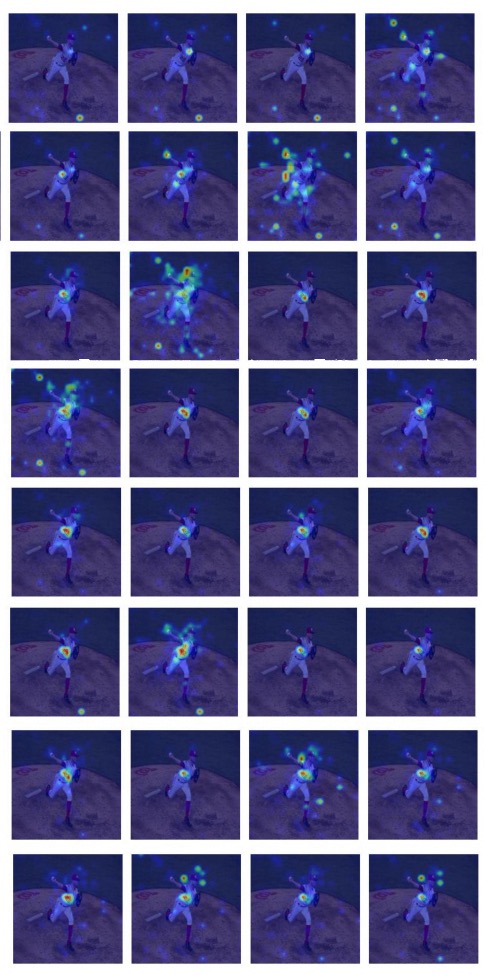}
    \caption{\textbf{Full details of the evolution of textual guidance across LLM layers.}}
    \label{fig:full_evolution}
\end{figure}

\section{Details of the Evolution of Textual Guidance}
In this part, we further analyze the evolution of textual guidance as the number of layers increases. In \cref{fig:full_evolution}, the question posed is "What number is on the player's jersey?". Our analysis reveals that the textual guidance does not immediately focus on the answer region but rather evolves systematically. Initially, the model focuses on the player and some irrelevant areas in the image. Subsequently, it shifts its attention to the entire body of the player, including the limbs and the clothing. The model then begins to concentrate on the number on the clothing. However, after initially focusing on the numbers, the model does not maintain a consistent focus; instead, it intermittently returns its attention to the player and the clothing. Starting from the 8th layer, the model consistently focuses on the number region, although there are still occasional shifts in attention to other areas. This process can be interpreted as the model continuously refining its reasoning and adjusting its focus, even revisiting and reconfirming previously identified regions, until it stabilizes on the correct answer.

\section{Imbalance Loss}
In this section, we further investigate the impact of balance loss on model performance. During our training process, under the constraint of balance loss, the model is required to select three different visual scales with as equal probability as possible. We believe that in real-world scenarios, different situations may have a preference for a particular visual scale. Therefore, by modifying the form of the balance loss, we constructed an imbalance loss to adjust the model's preferences. Specifically, we add coefficients to each scale, making the loss function as follows:
\begin{equation}
    L_{imbalance} = \alpha \sum_{i=0}^n w_i f_i P_i, 
\end{equation}
where \(w_i\) represents the penalty coefficient for the i-th branch. To align with the original loss values, \(w_i\) must satisfy \(\sum_{i=0}^n w_i = 3\). For the balance loss, \(w_i = 1\), meaning the coefficients for all branches are the same. Increasing \(w_i\) means increasing the penalty for that branch, leading the model to be less likely to choose it. Conversely, decreasing \(w_i\) makes the model more inclined to select that branch.
The results are shown in \cref{tab:imbalance}, where the list under "setting" in the table represents the penalty coefficients for the three branches, corresponding to 1x1 pooling, 2x2 pooling, and 4x4 pooling, respectively. We find that when the penalty coefficient for 1x1 pooling is increased, the model tends to perform more downsampling operations, which significantly degrades its performance on TextVQA. Conversely, decreasing the penalty coefficient for 1x1 pooling improves the model's performance on TextVQA. The ScienceQA test set is also affected, but with a completely opposite trend. We believe this is due to the differences in the tasks of the test sets. TextVQA requires the model to focus on text-related factual details in images, so a preference for finer scales is beneficial. In contrast, ScienceQA demands more reasoning from the model and does not require as much attention to the fine details of the image.
\begin{table}[h]
    \centering
    \caption{
    \textbf{Exploration of imbalance loss.}}
    \resizebox{0.8\linewidth}{!}{
        \begin{tabular}{lcccc}
            \toprule
            Setting             & TextVQA       & MME           & ScienceQA     & GQA           \\ \midrule
            Baseline            & \textbf{65.5} & 1562          & 76.4          & 61.7          \\
            {[}1.0,1.0,1.0{]}   & 65.2          & \textbf{1590} & 77.8          & \textbf{62.3} \\
            {[}1.1,1.0,0.9{]}   & 63.7          & 1581          & \textbf{78.0} & 61.1          \\
            {[}0.9, 1.0, 1.1{]} & 65.7          & 1587          & 76.1          & 62.1          \\ \bottomrule
        \end{tabular}
    }
    \label{tab:imbalance}
\end{table}

\end{document}